\pdfoutput=1
\documentclass[letterpaper, 10 pt, conference]{ieeeconf}
\IEEEoverridecommandlockouts
\usepackage[pdftex]{graphicx}
\usepackage{geometry}
\usepackage{stfloats}
\usepackage{amsmath,amssymb}
\usepackage{amstext}
\usepackage{mathtools}
\usepackage{bigints}
\usepackage{color}
\usepackage{xcolor}
\usepackage{cite}
\usepackage{blindtext}
\usepackage{etoolbox}

\usepackage{amsthm}

\usepackage[caption=false,font=footnotesize]{subfig}
\theoremstyle{remark}

\title{\LARGE \bf Anticipatory Human-Robot Path Planning for Search and Rescue}
\author{Barnabas~Gavin~Cangan, Larkin~Heintzman, Amanda~Hashimoto, Nicole~Abaid, and Ryan~K.~Williams
\thanks{
Barnabas~Gavin~Cangan, Larkin~Heintzman, Amanda~Hashimoto, Nicole~Abaid, and Ryan~K.~Williams are with the Department of Electrical and Computer Engineering, Virginia Polytechnic Institute and State University, Blacksburg, VA USA, 
\mbox{E-mail: \textit{\{bgavin, hlarkin3, ahashimo, nabaid, rywilli1\}@vt.edu}}
}%
\thanks{This work was supported by the National Science Foundation under grant CNS-1830414.}
}

\begin{document}

\definecolor{myOrange}{rgb}{1,0.5,0}
\newcommand {\rework}[1]{{\color{myOrange}#1}\normalfont}
\newcommand{\llh}[1]{{\color{blue}\textbf{llh: }#1}\normalfont}
\newcommand {\rkw}[1]{{\color{red}\textbf{Ryan: }#1}\normalfont}
\newcommand{\lbb}{\Big\{}
\newcommand{\rbb}{\Big\}}
\newcommand{\lb}{\left\{}
\newcommand{\rb}{\right\}}
\newcommand{\lp}{\Big(}
\newcommand{\rp}{\Big)}
\newcommand{\lpb}{\Big(}
\newcommand{\rpb}{\Big)}
\newcommand{\R}{\mathbb{R}}
\newcommand{\Q}{\mathbb{Q}}
\newcommand{\Z}{\mathbb{Z}}
\newcommand{\N}{\mathbb{N}}
\newcommand{\C}{\mathbb{C}}
\newcommand{\V}{\mathcal{V}}
\newcommand{\G}{\mathcal{G}}
\newcommand{\Op}{\mathcal{O}}
\newcommand{\M}{\mathcal{M}}
\newcommand{\E}{\mathcal{E}}
\newcommand{\A}{\mathcal{A}}
\newcommand{\Mt}{\mathcal{M}}
\newcommand{\I}{\mathcal{I}}
\newcommand{\mb}{\mathbf}
\newcommand{\vbar}{\,|\,}
\newcommand{\p}{\prime}
\newcommand{\eps}{\epsilon}
\newcommand{\dlt}{\delta}
\newcommand{\ol}{\overline}
\newcommand{\nth}{{\text{th}}}
\newcommand{\ttt}{\texttt}
\newcommand{\imp}{\Rightarrow}
\newcommand{\ifa}{\Leftrightarrow}
\newcommand{\hop}{\vspace{3mm}\hrule\vspace{3mm}\noindent}
\newcommand{\cont}{(\Rightarrow \Leftarrow)}

\maketitle

\begin{abstract}

In this work, our goal is to extend the existing search and rescue paradigm by allowing teams of autonomous unmanned aerial vehicles (UAVs) to collaborate effectively with human searchers on the ground. We derive a framework that includes a simulated lost person behavior model, as well as a human searcher behavior model that is informed by data collected from past search tasks. These models are used together to create a probabilistic heatmap of the lost person's position and anticipated searcher trajectories. We then use Gaussian processes with a Gibbs' kernel to accurately model a limited field-of-view (FOV) sensor, e.g., thermal cameras, from which we derive a risk metric that drives UAV path optimization. Our framework finally computes a set of search paths for a team of UAVs to autonomously complement human searchers' efforts.

\end{abstract}


\section{Overview} \label{sec:overview}

In 2018, over 650,000 lost persons were reported, and nearly 100,000 of these cases were answered with organized searches in either urban or wilderness environments by the search and rescue (SAR) community \cite{ncicreport}. Currently, search efforts usually involve a large team of personnel, in some cases over 100 people, working over several days to locate the lost person \cite{landSARcompat,ncicreport}. There have been many efforts to increase the effectiveness of SAR operations, taking into account recent advances in search theory \cite{landSARcompat} and informative path planning \cite{singh2009nonmyopic}. In addition, an unmanned aerial vehicle (UAV) recently was used in a SAR mission in Oregon that allowed searchers to visually inspect an area that otherwise would have been dangerous and time consuming to traverse \cite{van2017first}. However, the UAV  in \cite{van2017first} was a remotely controlled model, a Phantom 3 4K quadcopter manufactured by DJI, requiring a human pilot to provide continuous control and image processing.

In this work, we propose a framework to plan a set of paths for a team of UAVs to autonomously gather information about the environment and search for a lost person \cite{cangan2019risk}. To optimize measurements the UAVs take into account a lost person predictive model, topography in the search environment, and predicted human searcher trajectories. The lost person is modeled by taking into account prior beliefs as well as movement patterns informed from previous SAR missions \cite{hill1998psychology}. By utilizing the advantages afforded by UAVs, we aim to increase the \emph{efficiency} of SAR operations by minimizing the total time required while maximizing the chances of locating a lost person with available resources \cite{landSARcompat}.


\section{Modeling} \label{sec:modeling}

\begin{figure}[t]\centering
\includegraphics[width=0.65\linewidth]{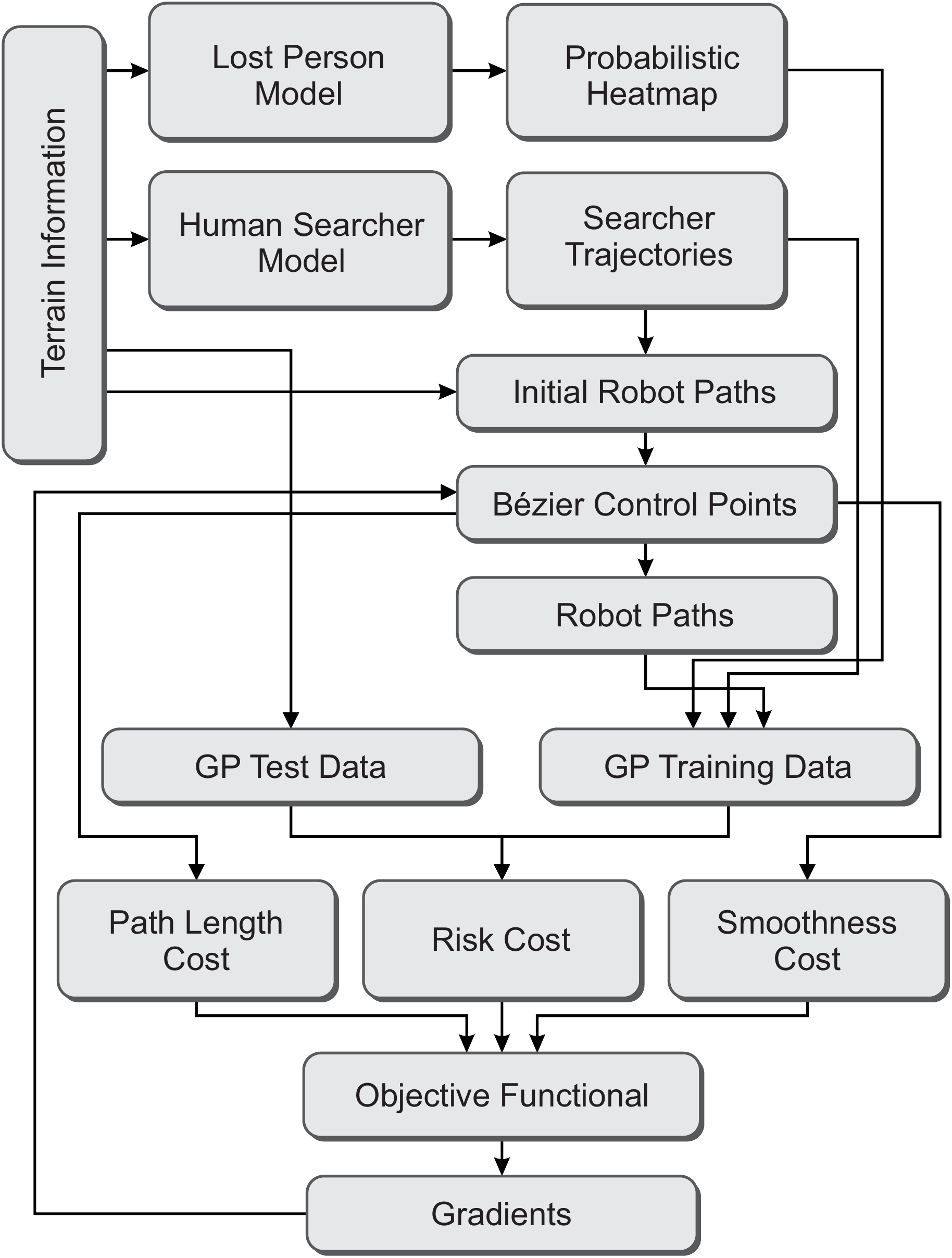}
\caption{Data flow diagram for the proposed human-robot planning framework.}
\label{fig:data_flow_diagram}
\end{figure}

In this Section we discuss the modeling of the lost person, the human searchers, UAV sensors, as well as the environment itself. For an overview of the data flow in the proposed framework, consider the diagram shown in Figure~\ref{fig:data_flow_diagram}.
\subsection{Human Searcher} \label{sub:human_searcher}
\begin{figure}[t]\centering
\subfloat[]{\includegraphics[width=0.45\linewidth]{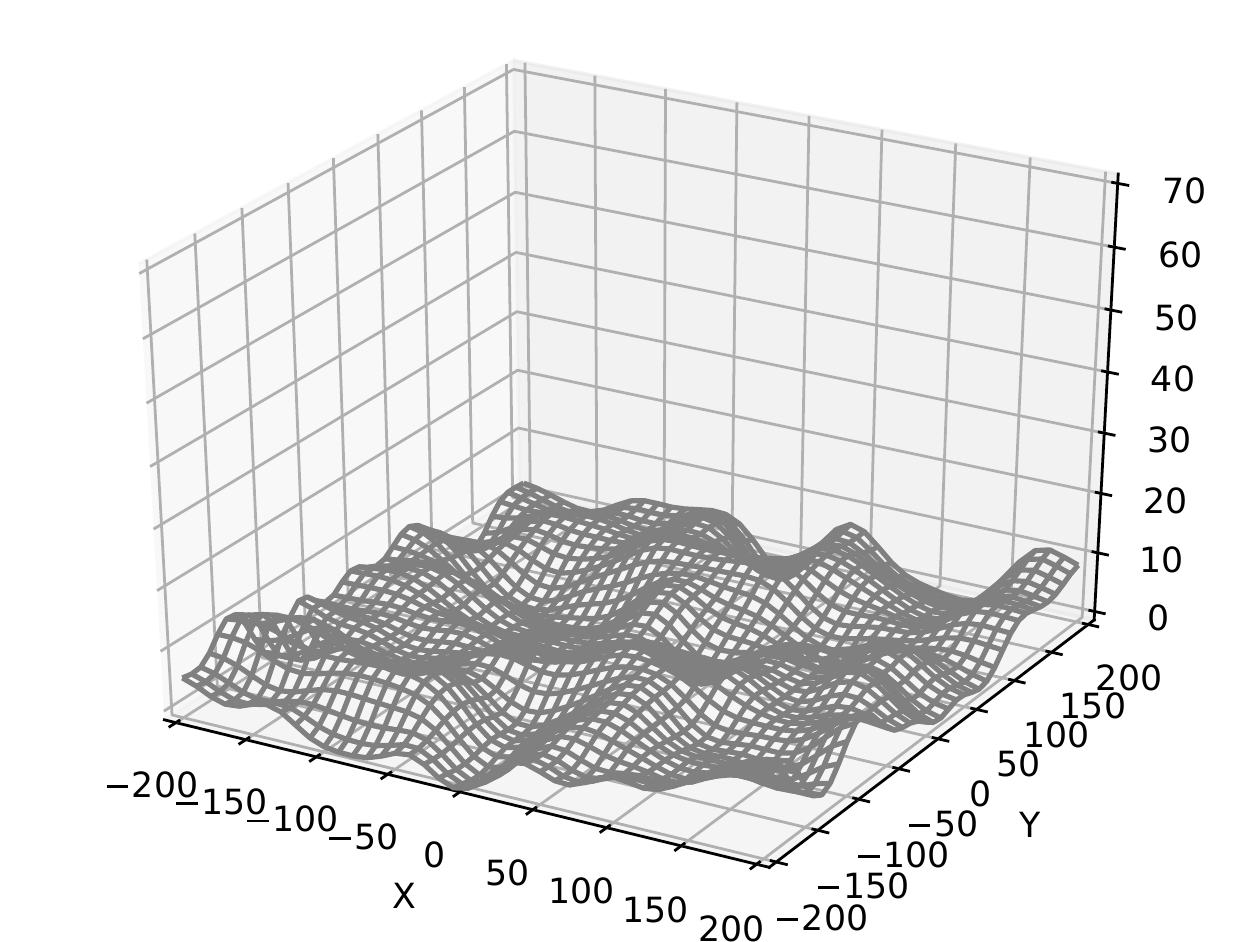}
\label{fig:ex_topo}}
\hfil
\subfloat[]{\includegraphics[width=0.45\linewidth]{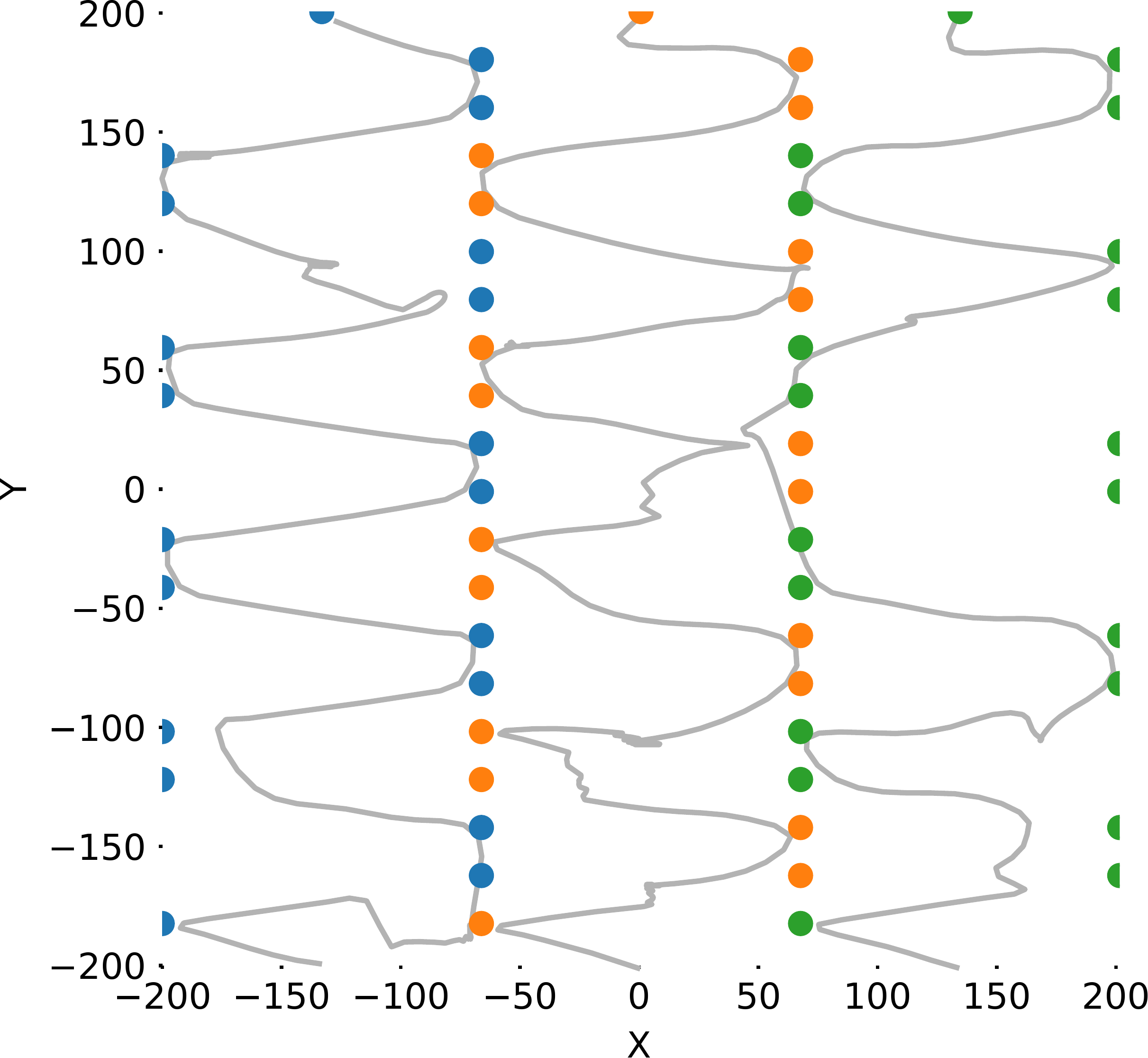}
\label{fig:ex_searcher_path}}
\caption{Showing in (a) simulated terrain and in (b) example anticipated searcher paths in grey, and the waypoints as colored dots.}
\label{fig:terrain_searcher_paths}
\end{figure}

In a land SAR mission, a particular area of land will have human searchers assigned to search it, which are referred to as \emph{sectors} \cite{koester2008lost}. We assume that each searcher has an entry and exit point for each sector, based on the overall sequence of sectors to be searched. The human searcher model we use here has two modes, a waypoint following mode and a gradient following mode. In the waypoint following mode, each searcher is represented as a self-propelled particle moving towards a predefined set of waypoints. Once a searcher is within a known radius of the current waypoint, the next waypoint becomes the next target. In our case, the waypoints are arranged to generate search paths, sometimes referred to as \emph{lawn mowing} paths. To better reflect reality, each searcher path is also influenced by the terrain gradient. For example, if the terrain in a sector becomes too steep to climb, the searcher is forced to navigate around the obstacle. In these cases, our human searcher model will switch to the gradient following mode of operation. However, there are cases where searchers may get stuck in steep trenches or canyons, and thus to combat this we included a tenacity parameter that gradually increases, allowing searchers to escape from these scenarios (i.e., local minima). Shown in Figure~\ref{fig:ex_topo} is an example of a simulated terrain map, and shown in Figure~\ref{fig:ex_searcher_path} is the resulting anticipated searcher paths using the terrain shown in Figure~\ref{fig:ex_topo}. Notice that the searchers' lawnmower paths are being influenced by the underlying terrain.
%
%
%
%
\subsection{Lost Person} \label{sub:lost_person}
People lost in the wilderness use one or more typical behaviors to move around, as seen in practice \cite{hashi2019agentbased,hill1998psychology}. These behaviors are labeled random traveling, route traveling, direction traveling, route sampling, direction sampling, view enhancing, backtracking, folk wisdom, staying put, and doing nothing \cite{hill1998psychology}. To simplify our lost person modeling, we will implement a subset of the above behaviors. The lost person motion model, coming from \cite{cangan2019risk}, used is:
\begin{equation} \label{eq:lost_person}
    m \ddot{x} + (a \| \dot x \| - b) \dot x = \alpha F^G (x_i) + \beta F^R
\end{equation}
Here we have chosen to model the lost person as an agent in an unknown environment with limited perception, with the assumption that environmental interactions and disturbances can be modeled as forces acting upon the agent. In \eqref{eq:lost_person}, $x$ is the agent position, $m$ is the inertia, $a$ is the friction, $b$ is the self-acceleration, $F^G(x_i)$ is an environmental force acting upon the agent which is a function of the current position, with $F^R$ as a stochastic process \cite{cangan2019risk}. As is common notation, $\alpha$ and $\beta$ are constant tuning parameters. Shown in Figure~\ref{fig:2d_paths}, under the searcher and UAV paths, is an example Monte Carlo simulation using this lost person model, along with a simulated topology shown in Figure~\ref{fig:ex_topo}, to generate a heatmap of possible locations. More accurate and complex lost person models are possible such as the one(s) described in \cite{hashi2019agentbased}. A benefit of the proposed work is that we can simply swap out the lost person model with an updated version without any modification to the current framework.
\subsection{UAV Measurement Model} \label{sub:uas_measurement}
\begin{figure}[t]\centering
\subfloat[]{\includegraphics[width=0.45\linewidth]{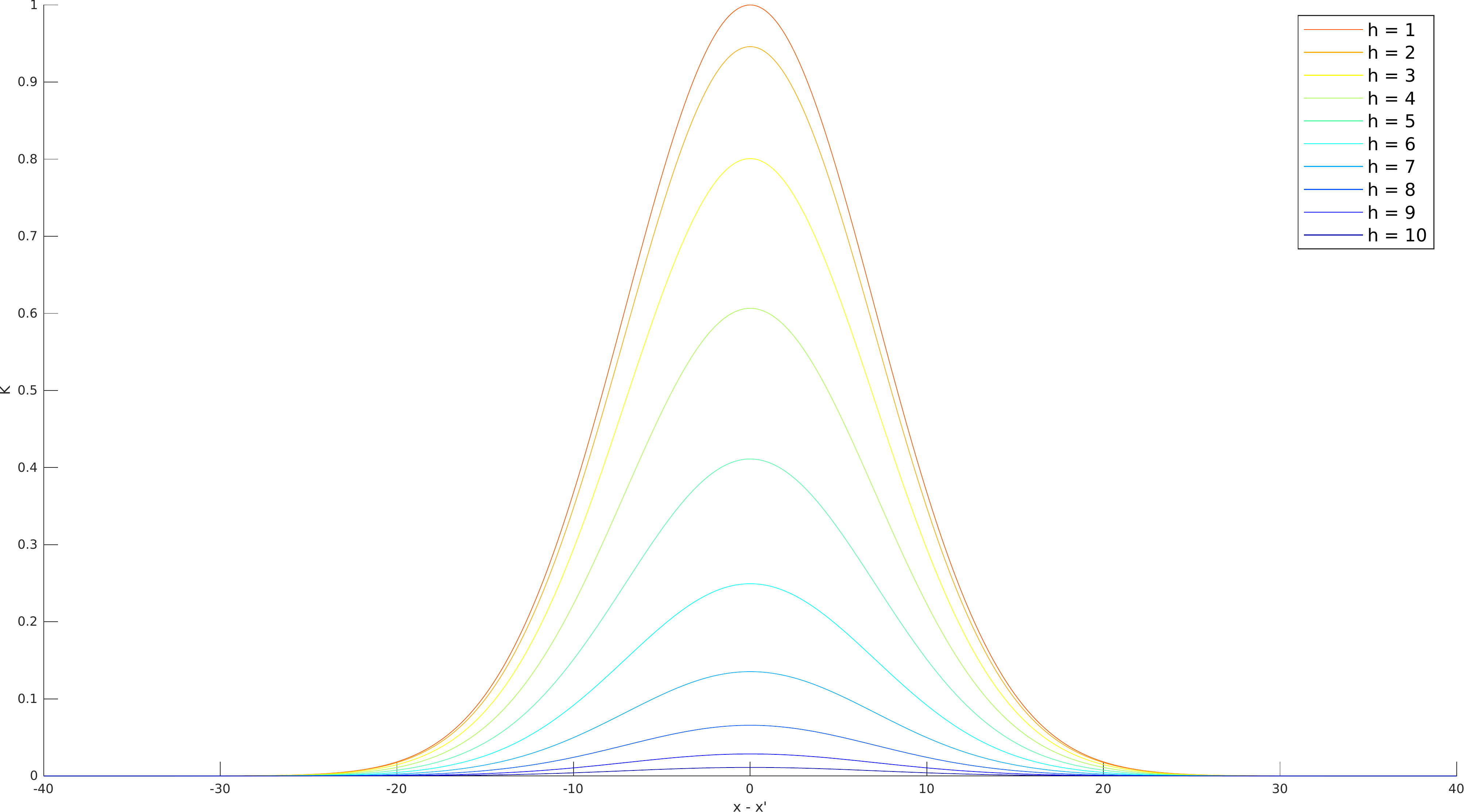}
\label{fig:ex_rbf_plot}}
\hfil
\subfloat[]{\includegraphics[width=0.45\linewidth]{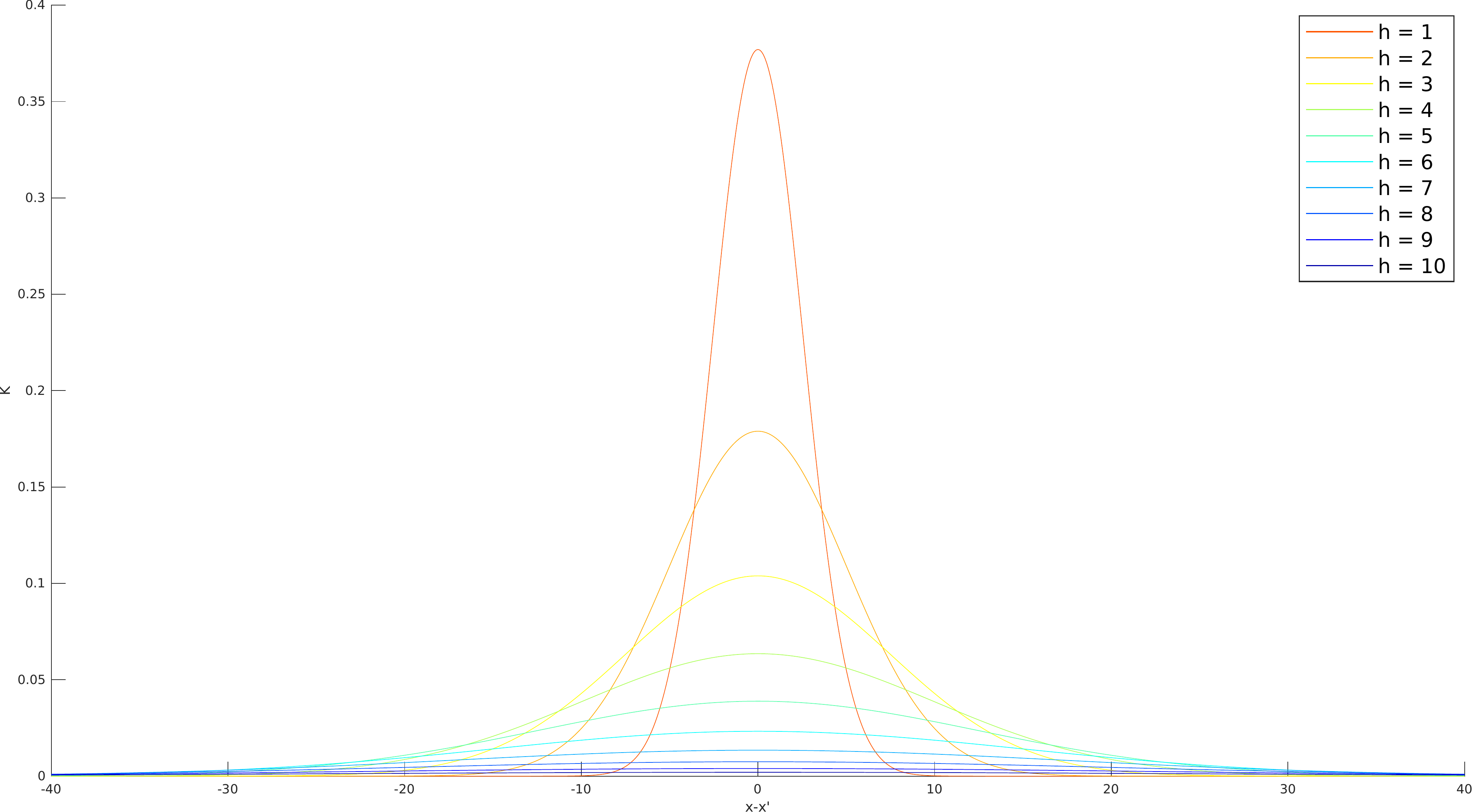}
\label{fig:ex_gibbs_plot}}
\caption{Showing in (a) a 2D measurement model that uses the RBF kernel, and in (b) a 2D measurement model that uses the Gibbs' kernel.}
\label{fig:kernel_comparison}
\end{figure}

We use a Gaussian process model (GPM) to combine the information from anticipated human searcher paths from Section~\ref{sub:human_searcher}, the lost person model from Section~\ref{sub:lost_person}, and UAV measurements. Using a GPM allows us to model realistic data collection from the UAVs via a Gibbs' kernel. This data collection represents a downward facing sensor such as a camera, LiDAR, or any other fixed FOV sensor that could be mounted on a UAV. With the proposed method, we can simulate data collection from any 3D point in the environment. We use a Gibbs kernel, as opposed to a radial basis function (RBF) kernel, because it allows for realistic modeling of \emph{fixed} FOV sensors. The RBF kernel does not capture the change in FOV with changing altitude, thus the Gibbs' kernel is better suited here. See Figure~\ref{fig:kernel_comparison} for a visual comparison between the RBF and Gibbs' kernels for the 2D case at various altitudes. Notice how the Gibbs' kernel captures the decrease in field of view we expect from a decreasing altitude.
\subsection{Initial Trajectory Sampling} \label{sub:initial_sampling}

Later we will describe our trajectory optimization approach, however a key part of that process is an initial set of trajectories to begin with. We get the initial set from the sampling based approach of RRTs \cite{cangan2019risk}. In our framework, we independently plan paths for all robots, using the same sector entry and exit points as the human searchers, and quantify the information gathered from all paths using the GPM. We consider the existence of no-fly zones in the environment which are represented as infinite height cylindrical obstacles. We do also take into account terrain elevation change in the trajectory planning to prevent collision.

\subsection{Risk Quantification} \label{sub:risk_quant}

\begin{figure}[t]\centering
\subfloat[]{\includegraphics[width=0.40\linewidth]{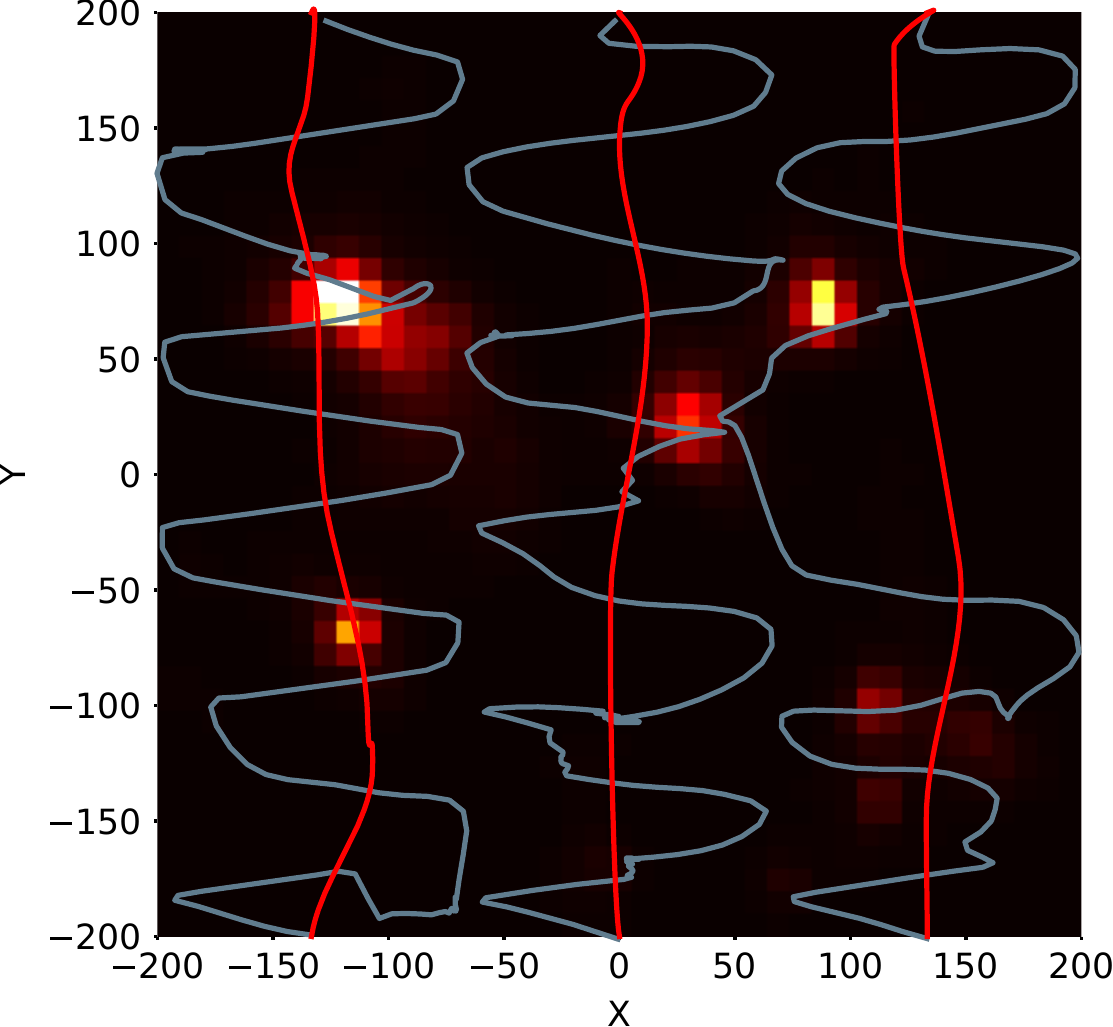}
\label{fig:2d_paths}}
\hfil
\subfloat[]{\includegraphics[width=0.5\linewidth]{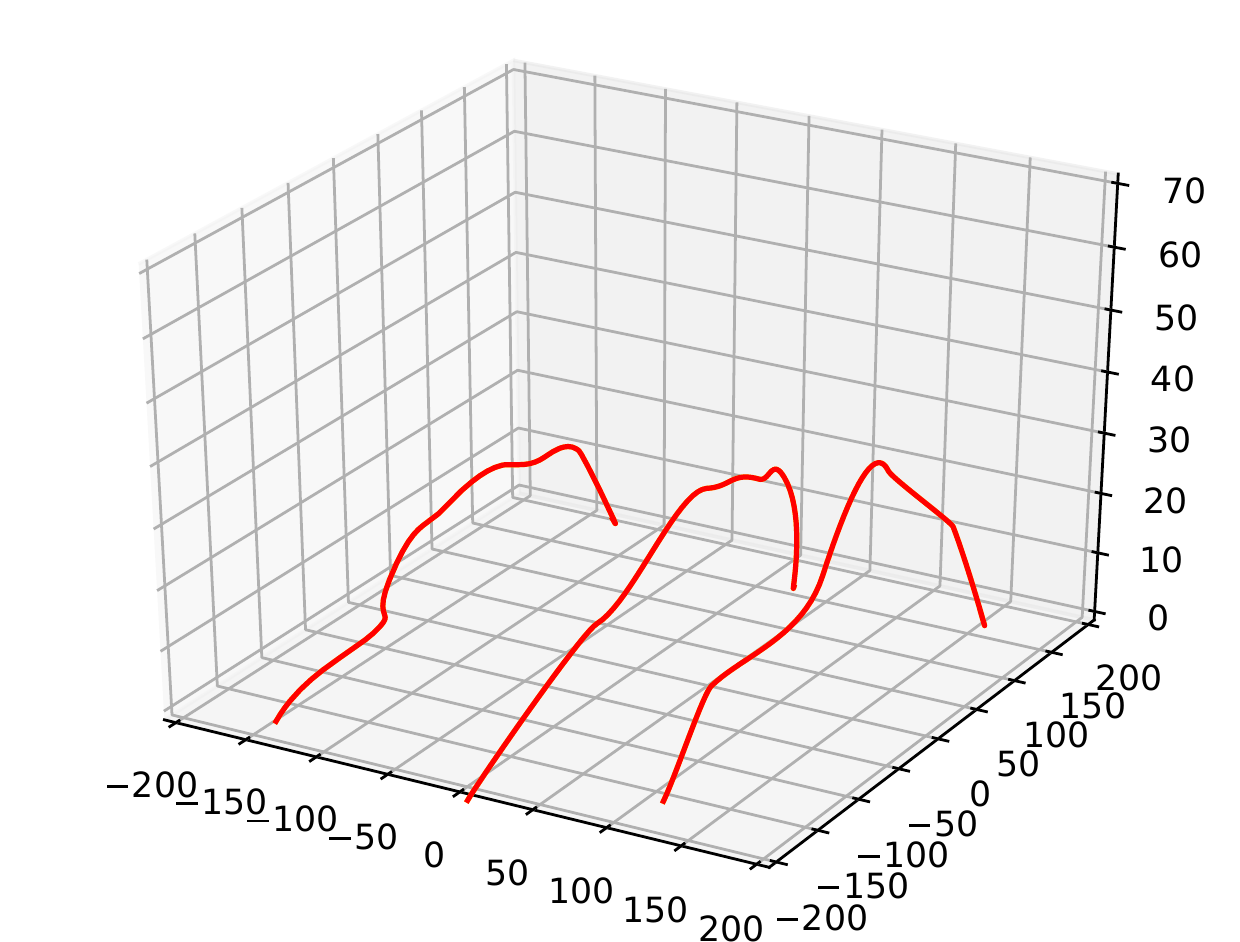}
\label{fig:3d_paths}}
\caption{Showing in (a) the lost person probability heatmap, with the searcher and UAV paths overlayed in blue and red, respectively. In (b) are the 3D versions of the UAV paths in (a).}
\label{fig:result_paths}
\end{figure}

In order to optimize paths for the UAVs, we need some quantification of the risk inherent in a trajectory based on the lost person model heatmap like the one with UAV paths shown in Figure~\ref{fig:2d_paths}, the anticipated human searcher trajectories like the ones shown in Figure~\ref{fig:ex_searcher_path}, and the set of assumed paths for the UAVs. In this case, the risk can also be restated as our \emph{belief} of the lost person's location with some uncertainty associated. We will omit a full derivation of the risk function for brevity, the details of which can be found in \cite{cangan2019risk}, and instead skip to the final result. Our risk function is given as:
\begin{equation} \label{eq:risk_function}
    \mathcal{R} = \sum_{i = 1}^n \frac{\sum X_i^*}{1 + \mu {\mathbf{X}_i^*}^2}
\end{equation}
where $X_i$ is a measurement at point $i$ indicating the presence of a lost person, $X^*_i$ is the \emph{prediction} from the GP at measurement point $i$, and $\mu$ a scaling term. Therefore risk is a function of the mean and variance associated with measurement locations given the data of the lost person model, the human searchers' anticipated paths, and the given set of UAV trajectories.

\subsection{Objective Function} \label{sub:objective_function}

As discussed in Section~\ref{sub:initial_sampling}, we have an initial set of trajectories that we must optimize to reduce the corresponding \emph{risk cost}, our system objective function. The risk cost gives us a measure of the uncertainty in the lost person's location, given the anticipated human searcher's paths and the given trajectories of the UAVs. The goal of the optimization process is to generate UAVs trajectories that effectively complement the ground searchers, subject to the given constraints. The constraints for us are start and end points, smoothness, and time expenditure which is expressed as a path length cap.

We parameterize the UAV paths with B\'ezier curves, which will help achieve the smoothness requirement while providing a convenient method of optimization. Using B\'ezier curves allows us to represent the risk cost as a function of the sparse set of parameters $\lambda \in \mathbb{R}^{p \times D}$, where $p \in \mathcal{N}$, which define trajectories given by $\theta_{\lambda}(t)$. Formally, our goal is to find a set of parameters, $\lambda$, to minimize the risk metric (risk cost), that is:
\begin{equation} \label{eq:argmin_risk}
    \lambda^* = \underset{\lambda \in \Lambda}{\arg \min}\; \mathcal{R}\left[ \theta_{\lambda}(t) \right]
\end{equation}
The resulting set of UAV trajectories would then be $\theta_{\lambda^*}(t)$. We can compute the total time expense of the path by approximating it's total length via the second derivative of the cubic B\'ezier curve see \cite{cangan2019risk} and \cite{vincent2001fast} for details. We can similarly compute the smoothness of a set of trajectories by summing over the square differences of parameter derivatives. Using path length and path smoothness constraints, we can restate our goal as:
\begin{equation} \label{eq:objective_function}
\begin{aligned}
    &\text{minimize } \mathcal{R} \left[ \theta_{\lambda}(t) \right]\\
    &\quad\text{subject to } \mathcal{L} \left[ \theta_{\lambda}(t) \right] \leq \mathcal{C}_{\text{time}}\\
    &\quad\text{and } \mathcal{S} \left[ \theta_{\lambda}(t) \right] \leq \mathcal{C}_{\text{smooth}}
\end{aligned}
\end{equation}
where $\mathcal{L} \left[ \theta_{\lambda}(t) \right]$ is the total path length cost, and $\mathcal{S} \left[ \theta_{\lambda}(t) \right]$ is the total path smoothness cost, associated with trajectories $\theta_{\lambda}(t)$. We can state \eqref{eq:objective_function} more cleanly by combining each term into an \emph{objective functional} given as:
\begin{equation} \label{eq:combined_objective}
    \mathcal{F} \left[ \theta_{\lambda}(t) \right] = \mathcal{R} \left[\theta_{\lambda}(t) \right] + \alpha_{\mathcal{L}} \mathcal{L} \left[\theta_{\lambda}(t) \right] + \alpha_{\mathcal{S}} \mathcal{S} \left[\theta_{\lambda}(t) \right]
\end{equation}
where $\alpha_{\mathcal{L}}$ and $\alpha_{\mathcal{S}}$ are scaling parameters for the path length constraint and the path smoothness constraint, respectively. The restated goal is to minimize the objective functional $\mathcal{F} \left[ \theta_{\lambda}(t) \right]$ to maximize the efficiency of the planned UAV trajectories.

\subsubsection{Trajectory Optimization}

Here we discuss the gradient descent process of trajectory optimization for the objective function in \eqref{eq:combined_objective}. Specifically, we use an iterative approach, where in each iteration the gradient is computed, $\nabla_{\lambda} \mathcal{F} \left[ \theta_{\lambda} \right]$, about the current trajectory set, $\theta_{\lambda}$, with respect to the current parameter set $\lambda$.  The parameter set is then propagated at each iteration by following the direction of steepest descent as defined by the gradient of the objective functional:
\begin{equation} \label{eq:gradient_descent}
    \lambda_{i + 1} = \lambda_i - \eta \nabla_{\lambda} \mathcal{F} \left[ \theta_{\lambda} \right],\quad i = 1, 2, \dots
\end{equation}
where $\eta$ is a tunable descent rate parameter. Since the objective functional is of large dimension and highly non-convex, we use an optimization method typically found in neural network-based machine learning applications, referred to as \emph{Adam} \cite{kingma2014adam}. Adam features adaptive learning rate adjustment based on first and second moments of the gradient \cite{kingma2014adam}.

Even using Adam, the optimization process can be quite slow and thus we employ a technique to improve computation time \cite{cangan2019risk}. In the inference step, as part of the GPM, points from the grid are placed in a square covariance matrix which can be slow to compute. To combat this slowdown, we sort the rows of this matrix based on the Morton Z-order space filling curve \cite{morton1966computer}. Using this technique allows us to matrix position to spatial location, meaning we can approximate large blocks of the matrix as zeros as they correspond to spatially distant points.

\section{Results} \label{sec:results}

Here we discuss our implementation and present several simulations to demonstrate the effectiveness of the proposed framework.

\subsection{Implementation} \label{sub:implementation}

The components of our framework, lost person model, human searcher model, sampling-based planner, 3D cubic B\'ezier curve fitting, GP inference, and gradient-based optimization were implemented in Python v3.6.8 \cite{cangan2019risk}. NumPy was used for numerical computations, and a custom GP package was built using PyTorch so that we could guarantee support for a 3D Gibbs' kernel.

\subsection{Simulation} \label{sub:simulations}

Here we present simulations on the proposed framework and implementation. For the coming simulated experiments, the terrain is randomly generated on an area representing 40 acres ($400\text{m} \times 400\text{m}$). The lost person model \eqref{eq:lost_person} was built with the following values: $m = 70$, $a = 10^{−3}$, $b = 10^{−5}$, $\alpha = −5$, and $\beta = 0.3$. The lost person model Monte Carlo simulation used 25000 iterations in total, and the terrain resolution was $10\text{m} \times 10\text{m}$.
\subsection{Qualitative Results} \label{sub:qualitative_results}
Here we show the UAV paths that result from using the proposed framework, we provide these plots for intuition and visualization of factors that affect these trajectories. Shown in Figure~\ref{fig:result_paths} is a top view of the UAV paths generated, for the terrain and searcher paths shown in Figures~\ref{fig:ex_topo} and \ref{fig:ex_searcher_path}.

\begin{figure}[t]\centering
\subfloat[]{\includegraphics[width=0.4\linewidth]{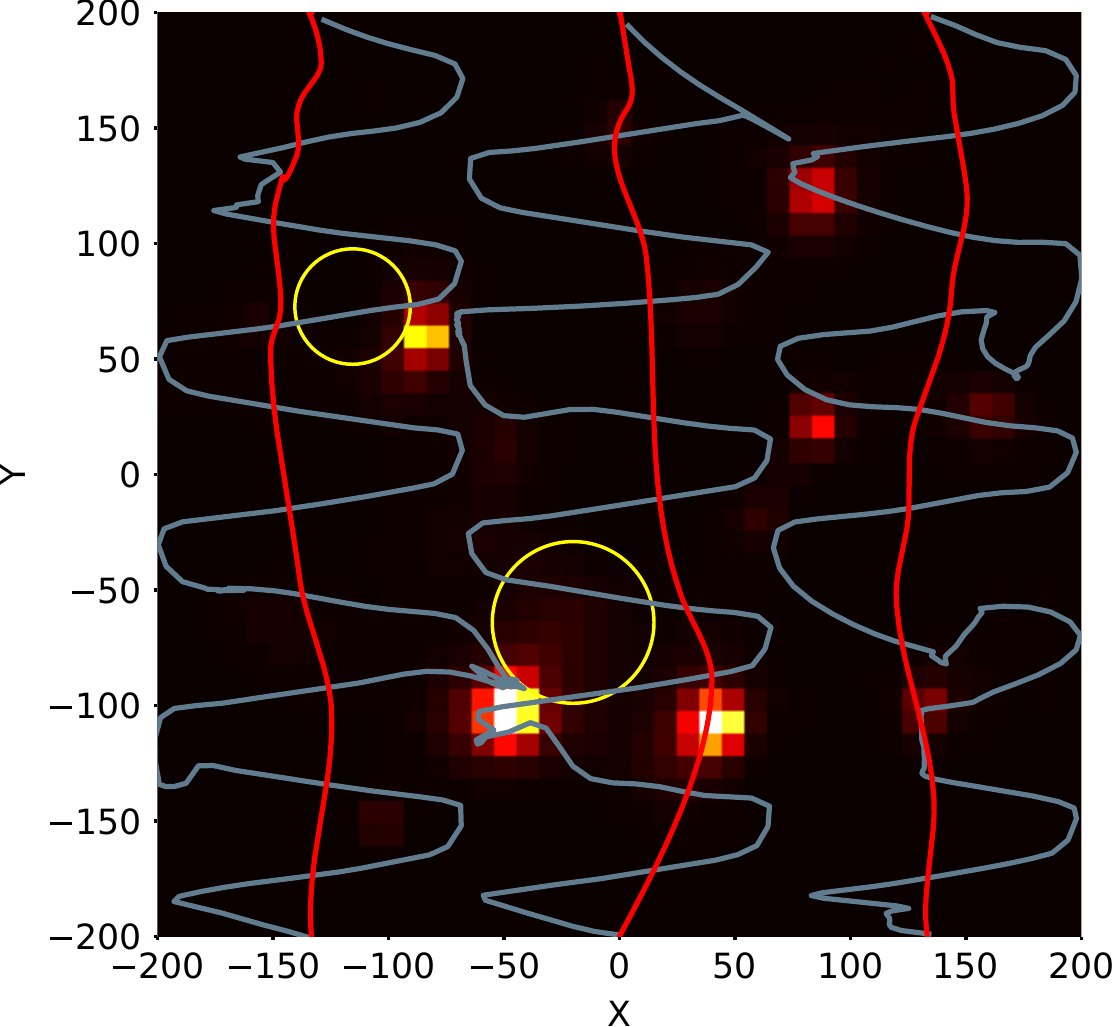}
\label{fig:2d_paths_obstacle}}
\hfil
\subfloat[]{\includegraphics[width=0.5\linewidth]{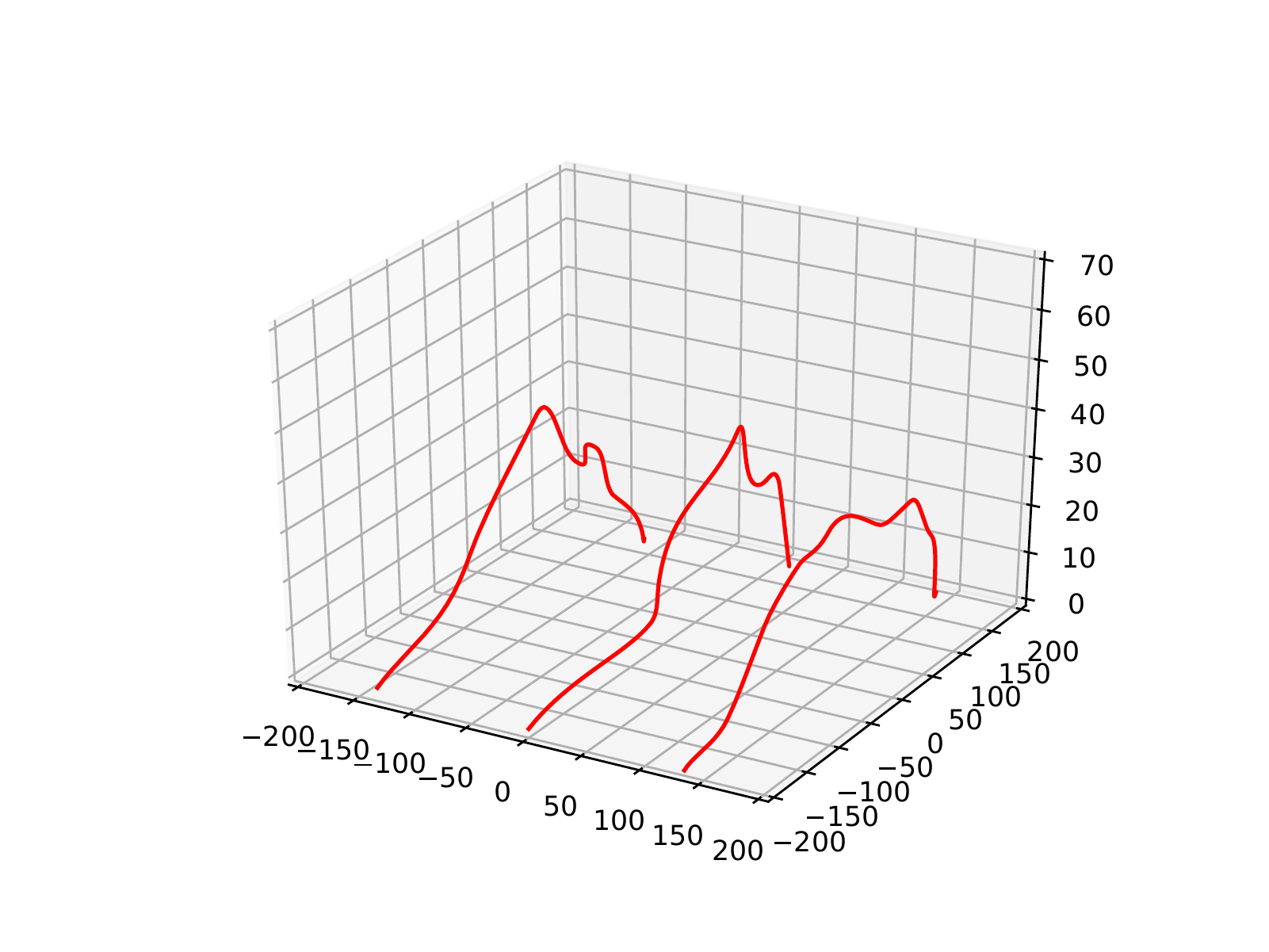}
\label{fig:3d_paths_obstacle}}
\caption{Showing in (a) the lost person probability heatmap, with the searcher and UAS paths overlayed in blue and red, respectively. The no-fly zones are represented as yellow circles. In (b) are the 3D versions of the UAS paths in (a).}
\label{fig:result_paths_obstacle}
\end{figure}

Experiments were also done with \emph{no-fly} zones in the environment, which are represented as infinitely tall cylindrical obstacles. We use a delayed collision checking approach wherein collision is checked for paths that have a lower risk cost than the last known set of minimal risk paths. Shown in Figure~\ref{fig:result_paths_obstacle} are the results of experiments with no-fly zones included in the environment, here shown as yellow circles.

As shown in Figures~\ref{fig:result_paths} and \ref{fig:result_paths_obstacle}, the proposed framework is able to incorporate information from multiple sources to compliment human searcher efforts. The sources are the lost person model Monte Carlo simulation, the anticipated human searchers' paths, and the terrain. The UAV paths are clearly drawn towards those areas not already covered by human searchers, and among those especially towards the high probability points of the lost person simulation. Figures~\ref{fig:3d_paths} and \ref{fig:3d_paths_obstacle} show that the framework can effectively plan in $3$D using the altitude based FOV measurement model, balancing between high and low altitude flight to minimize risk.

Since the framework searches locally to improve paths, it is possible we settle into a local minimum of the risk cost. However, local minima are a common problem for iterative optimization schemes, and there are methods to address this issue, such as using stochastic gradient descent \cite{needell2016stochastic}. We have not implemented any method to address local minima in this work, but we note that it would not be a major extension to do so.

\subsection{Quantitative Results} \label{sub:quantitative_results}

For a quantitative evaluation of the proposed framework, we compare against three cases that relate to how SAR operations function: a) human searchers performing a search task without the assistance of UAVs, the most common circumstance currently \cite{van2017first}, b) human searchers performing a search with manual UAVs that follow the same path as the human searchers at a fixed height of $15$m over the terrain, c) human searchers with fully autonomous UAVs that follow the shortest collision-free path, as computed by RRT$^*$. See Table~\ref{tab:quant_results} for the results from these comparisons, where a lower risk cost is desirable, and the rows correspond to cases a), b), and c) respectively.
\begin{table}[ht]
    \centering
    \caption{\label{tab:quant_results}Quantitative comparison results.}
    \begin{tabular}{|c|c|c|c|}
        \hline
        Scenario & Risk Cost $(\times 10^{18})$ & \% of Max & Planning Time\\
        \hline
        No UAVs & 4.846 & 100 & N/A\\
        \hline
        UAVs, RRT$^*$ & 4.412 & 91.044 & 20.397s \\
        \hline
        UAVs, Manual & 4.122 & 85.056 & N/A \\
        \hline
        UAVs, Risk & 3.389 & 69.937 & 1193.781 \\
        \hline
    \end{tabular}
\end{table}
From the results in Table~\ref{tab:quant_results}, clearly the human searchers benefit, in terms of risk, from using even manually controlled UAVs because of the high altitude coverage advantage they offer. The risk cost in the manually controlled case is lower than the shortest path RRT$^*$ case because the paths planned via RRT$^*$ are usually close to straight lines in the absence of obstacles, and they are at a lower height than in the manual control case, resulting in more risk overall. Further, the proposed method, wherein UAVs plan paths to explicitly minimize risk, performs better than all others by complementing searchers' efforts and controlling altitude effectively, balancing FOV and quality of measurement.

The scenarios considered here are reasonably consistent with how searches are currently performed within the SAR community \cite{van2017first}, and the results in Table~\ref{tab:quant_results} indicate that using autonomous aerial agents, here modeled as UAVs, can greatly improve the success of the search mission via minimizing risk.

\section{Conclusions} \label{sec:conclusions}
In this extended abstract we have presented a framework for allowing unmanned aerial vehicles to assist SAR personnel in search tasks.


\end{document}